\documentclass[sigconf,screen=true,natbib=false,nonacm]{acmart}
\usepackage{amsmath,amsfonts}
\usepackage{multirow}
\usepackage{multicol}
\usepackage{amsthm}
\usepackage{listings}
\usepackage{algorithm}
\usepackage{algpseudocode}
\usepackage{graphicx}
\usepackage{textcomp}
\usepackage{xcolor}
\usepackage{cleveref}
\usepackage{booktabs}
\usepackage{comment}
\usepackage{tabularx}
\usepackage{multirow}
\usepackage{bm}
\usepackage{url}
\usepackage{xspace}
\usepackage{pifont}
\usepackage{arydshln}
\usepackage{verbatim}
\usepackage[referable]{threeparttablex}
\usepackage{calc} 
\usepackage{color}
\usepackage{float}
\usepackage{subcaption}
\usepackage[utf8]{inputenc}
\usepackage{booktabs} 
\usepackage{tabularx} 
\usepackage{tikz}
\usetikzlibrary{shapes.geometric, arrows}
\tikzstyle{startstop} = [rectangle, rounded corners, minimum width=3cm, minimum height=1cm,text centered, draw=black]
\tikzstyle{io} = [trapezium, trapezium left angle=70, trapezium right angle=110, minimum width=3cm, minimum height=1cm, text centered, draw=black]
\tikzstyle{process} = [rectangle, minimum width=3cm, minimum height=1cm, text centered, draw=black]
\tikzstyle{arrow} = [thick,->,>=stealth]

\usepackage{filecontents}                                  
\usepackage{pgfplots}
\usepackage{pgfplotstable}
\usepackage{scalefnt}
\pgfplotsset{compat=newest}

\newcommand{\tensor}[1]{\mathcal{#1}}      
\renewcommand{\vec}[1]{\boldsymbol{#1}}    

\definecolor{USTgold}{RGB}{153,102,0}
\definecolor{USTyellow}{RGB}{204,153,0}
\definecolor{USTyellowlight}{RGB}{255,212,0}
\definecolor{USTorange}{RGB}{255,166,26}
\definecolor{USTpink}{RGB}{255,157,157}
\definecolor{USTblue}{RGB}{0,51,102}
\definecolor{USTmiddle}{RGB}{0,116,188}
\definecolor{USTlight}{RGB}{99,202,225}
\definecolor{USTgray}{RGB}{204,204,204}
\definecolor{USTred}{RGB}{237,27,47}
\definecolor{USTdarkred}{RGB}{124,35,72}

\definecolor{CUHKorange}{RGB}{244,106,18} 
\definecolor{CUHKblue}{RGB}{0,111,190}    
\definecolor{CUHKgreen}{RGB}{0,127,128}   
\definecolor{CUHKred}{RGB}{228,46,36}     
\definecolor{CUHKyellow}{RGB}{198,148,34} 
\definecolor{CUHKdark}{RGB}{114,44,114}   
\definecolor{CUHKmiddle}{RGB}{144,44,144} 
\definecolor{CUHKlight}{RGB}{167,44,167}

\iftrue
\def\BibTeX{{\rm B\kern-.05em{\sc i\kern-.025em b}\kern-.08em
    T\kern-.1667em\lower.7ex\hbox{E}\kern-.125emX}}

\fi

\acmConference[DAC '24]{Design Automation Conference}{June 23--27, 2024}{San Francisco, CA}

\newcommand{\minisection}[1]{\vspace{.06in}\noindent{\textbf{#1}}}

\iftrue
\fi

\iftrue
\usepackage{titlesec}
\titlespacing\section{2pt}{5pt plus 1pt minus 1pt}{0pt plus 1pt minus 1pt}
\titlespacing\subsection{2pt}{5pt plus 1pt minus 1pt}{0pt plus 1pt minus 1pt}
\titlespacing\subsubsection{2pt}{5pt plus 1pt minus 1pt}{2pt plus 1pt minus 1pt}
\usepackage[inline]{enumitem}
\setlist{leftmargin=5.08mm}
\fi

\iftrue
\fi

\newcommand{\etal}{\textit{et al}.}

\begin{document}
\title{CAMO: Correlation-Aware Mask Optimization with Modulated Reinforcement Learning}

\iftrue
\author{Xiaoxiao Liang}
\affiliation{HKUST(GZ)}

\author{Haoyu Yang}
\affiliation{NVIDIA}

\author{Kang Liu}
\affiliation{HUST}

\author{Bei Yu}
\affiliation{CUHK}

\author{Yuzhe Ma}
\affiliation{HKUST(GZ)}
\fi

\begin{abstract}
Optical proximity correction (OPC) is a vital step to ensure printability in modern VLSI manufacturing. 
Various OPC approaches based on machine learning have been proposed to pursue performance and efficiency, which are typically data-driven and hardly involve any particular considerations of the OPC problem, leading to potential performance or efficiency bottlenecks. 
In this paper, we propose CAMO, a reinforcement learning-based OPC system that specifically integrates important principles of the OPC problem.
CAMO explicitly involves the spatial correlation among the movements of neighboring segments and an OPC-inspired modulation for movement action selection.
Experiments are conducted on both via layer patterns and metal layer patterns. 
The results demonstrate that CAMO outperforms state-of-the-art OPC engines from both academia and industry.
\end{abstract}

\maketitle
\pagestyle{empty}

\section{Introduction}
\label{section1}
The scaling down of the modern VLSI system has aggravated the diffraction effects in manufacturing and greatly challenged the yield. Optical proximity correction (OPC) aims to compensate for the lithography proximity effects by making corrections to the patterns on a mask. 
It has been a long-term critical problem in chip manufacturing. 
Therefore, various solutions have been proposed from both industry and academia \cite{OPC-SPIE1994-Otto,OPC-DATE2015-Kuang,OPC-TCAD2016-Su,OPC-ICCAD2014-Awad,OPC-TIP2007-Poonawala,OPC-DAC2014-Gao}, which leveraged empirical experience, iterative local search, or numerical optimization to generate the final mask. 
These methods have become the foundation for modern mainstream OPC flow in commercial tools. 
However, due to the inherent complexity of the lithography process, achieving the desired solution remains challenging.

Recently, the rise of machine learning (ML) techniques has provided new opportunities and solutions for OPC problems. 
Existing ML-based OPC techniques could be broadly classified into three categories: regression models~\cite{OPC-SPIE2015-Matsunawa,OPC-TSM2008-Gu}, generative OPC models~\cite{OPC-TCAD2020-Yang, OPC-ICCAD2020-Chen,OPC-TCAD2021-Shao}, and reinforcement learning (RL)-based OPC~\cite{RLOPC-TCAD2023-Liang}.
Note that although A2ILT~\cite{OPC-DAC2022-Wang} also adopts RL in the OPC problem, essentially it belongs to the inverse lithography technique but leverages RL to guide the gradient calculation. 
Generative OPC mainly utilizes generative models such as generative adversarial networks (GANs). 
For instance, conditional GANs have been extensively used in previous works~\cite{OPC-TCAD2020-Yang,OPC-ICCAD2020-Chen,OPC-TCAD2021-Shao}, in which the OPC problem is treated as an image generation task.
The regression-based and RL-based OPC share a similar principle, where a model is trained to determine the movement of the edge segments in a pattern.
The regression-based OPC methods~\cite{OPC-TSM2008-Gu,OPC-SPIE2015-Matsunawa} train the model purely based on the supervision data, while the RL-based OPC~\cite{RLOPC-TCAD2023-Liang} maintains an agent in the form of a neural network, which analyzes the layout geometry and determines how the patterns should be corrected based on the obtained reward.

\begin{figure}[!tb]
	\centering
	\includegraphics[width=.886\linewidth]{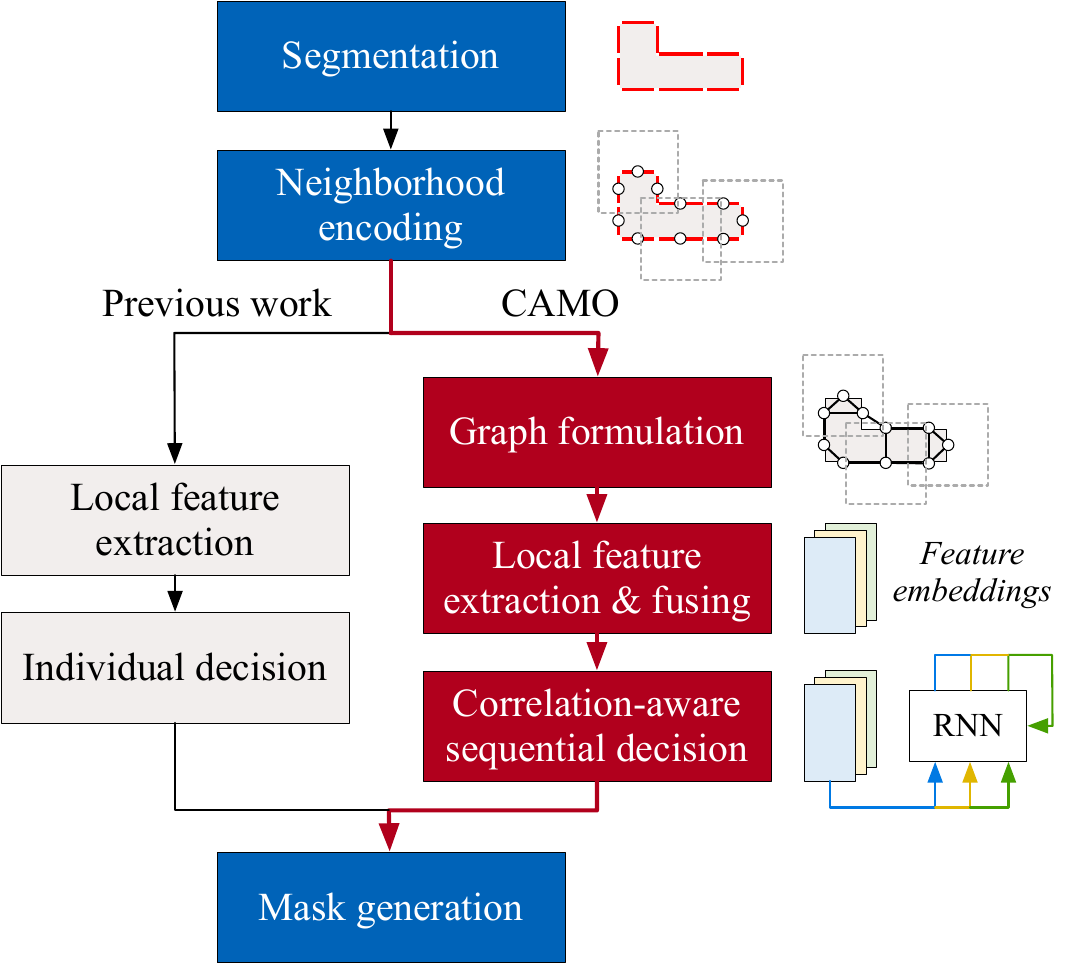}
    \caption{The comparison of the inference flow among CAMO and previous regression-based and RL-based OPC. }
 \label{fig:intro1}
 \end{figure}


However, there are several issues with the aforementioned ML-OPC techniques. 
Generative learning-based~\cite{OPC-ICCAD2020-Chen} and regression-based OPC approaches~\cite{OPC-TSM2008-Gu,OPC-SPIE2015-Matsunawa} utilize a given dataset that contains designs and optimized masks from other OPC engines as the supervision, hence their performance may be bounded by the quality of the dataset~\cite{RLOPC-TCAD2023-Liang}. 
Although RL-based OPC can intrinsically perform exploration to search for optimal masks, the training is usually very inefficient due to the huge action space. 
Besides, previous ML-based OPC approaches are typically data-driven, while some OPC-specific principles are hardly involved in the methods, which may lead to potential bottlenecks in performance or efficiency. 


The lithography principles suggest that the closely located segments are spatially \textit{correlated}, i.e., the printed contour in a region is determined by the movements of a set of neighborhood segments. 
In regression-based OPC~\cite{OPC-TSM2008-Gu,OPC-SPIE2015-Matsunawa} and RL-based OPC~\cite{RLOPC-TCAD2023-Liang}, the prediction for each possible movement depends solely on the segment's local features, while lacking coordination among the actual movements of the neighborhood segments, which leads to fluctuation of the mask quality.
Similar concerns regarding the spatial correlation in OPC were mentioned in Awad~\etal~\cite{OPC-ICCAD2014-Awad}, which models the regional light intensity with respect to adjacent segment pairs, such that the adjacent segments' movement can be coordinated. 
%

In this paper, we propose a spatial correlation-aware OPC framework, namely CAMO, that processes multiple segments by capturing their local correlation. 
CAMO performs OPC by moving the edge segments and adopting a policy gradient-based RL paradigm.  
Moreover, several attempts have been made in CAMO to capture the spatial correlation in the OPC problem. 
Firstly, different from RL-OPC~\cite{RLOPC-TCAD2023-Liang}, CAMO encodes the layout information into a graph, where each graph node represents a segment on the pattern boundary, and the graph edge is determined by the spatial proximity between the segments. 
CAMO then utilizes a graph neural network (GNN) for node embedding generation, which allows the node features to fuse along the graph edges and captures more intensive local information.
Secondly, taking advantage of a recurrent neural network (RNN) in sequential modeling, we employ an RNN to sequentially analyze the node embeddings and make decisions for segment movement on the fly, such that the spatial correlation can be captured and the movements among neighborhood segments can be coordinated to achieve superior mask quality. 

Considering the potential huge solution space of the OPC problem, a purely data-driven learning scheme may not be efficient enough to scale to layers with complex patterns, e.g., metal layers, and hence many previous works fail to demonstrate the effectiveness on those layers \cite{OPC-ICCAD2020-Chen,RLOPC-TCAD2023-Liang}.
In CAMO, we address the efficiency and generalization issues by incorporating the domain knowledge of OPC.
More specifically, an OPC-inspired \textit{modulator} is proposed to modulate the RL agent output and boost the optimization. 
Therefore, our RL model can be trained much more easily while generating desired masks. 
The technical contributions of this paper can be summarized as follows:
\begin{itemize}
    \item An RL-based OPC framework is built for modern mask optimization;
    \item The spatial correlation is well-considered, including a GNN-based feature fusion and an RNN module in the policy structure to capture the spatial correlation when handling multiple segments;
    \item We propose an OPC-inspired modulation module to guide the training in RL to achieve more stable and more efficient training;
    \item Experiments are conducted on both via layer patterns and metal layer patterns, where CAMO outperforms state-of-the-art OPC techniques from academia and industry commercial tools.
\end{itemize}
\section{Preliminaries}

\subsection{Reinforcement Learning}
Reinforcement learning (RL) is a paradigm of how intelligent agents optimize a defined objective function by engaging in a sequence of interactions with their environment. 
Specifically, the interaction sequence involves a set of optimization instances denoted as \textit{state} $s$. In each step, with a decision model namely policy $\pi$ maintained, the agent analyzes the current state $s$ and selects over a set of available actions $A$. 
Triggered by the selected action $a$, the state then transits to the next state $s'$, on which the environment performs evaluation with the \textit{reward} $r(s, a)$ generated. 
The decision and environmental interaction process is iteratively performed, which forms a Markov chain:
\begin{equation}
    {s_0}\stackrel{a_0}{\longrightarrow}({s_1},r_1)\stackrel{a_1}{\longrightarrow}...\stackrel{a_{T-1}}{\longrightarrow}({s_T},r_T).
    \label{eqn:Markov}
\end{equation}
The continuous update ends when the final state is reached or the maximum number of steps allowed $T$ is consumed.

The objective of the RL process is to maximize the accumulative reward over the state-updating trajectory, formulated as follows:
\begin{equation}
\begin{split}
    \mathop {\max }\limits_\theta  J(\theta ) &=   \sum\nolimits_\tau  {R(\tau ){p}(\tau|\theta )},\\
    R(\tau ) &= \sum\nolimits_t {\gamma^{t-1}{r_t}},
\end{split}
\label{eqn:max-rwd}
\end{equation}
where $\tau$ denotes a trajectory as depicted in \Cref{eqn:Markov}, $p(\tau|\theta)$ denotes the probability that $\tau$ occurs when the policy is parameterized by $\theta$, and $\gamma$ is the discount factor.

\subsection{Policy Gradient}
Policy gradient is an RL algorithm that tunes the policy parameters to increase the likelihood of taking actions that result in better accumulative rewards~\cite{REINFORCE}. 
In the implementation, the gradient of the expected accumulative reward $J(\theta)$ w.r.t. the policy parameters $\theta$ is computed, which then serves as an estimate of the policy parameters updating direction. 


\section{Method}
\subsection{RL Environment}
\label{sec:RLEnv}
\begin{figure*}[!tb]
	\centering
	\includegraphics[width=.86\linewidth]{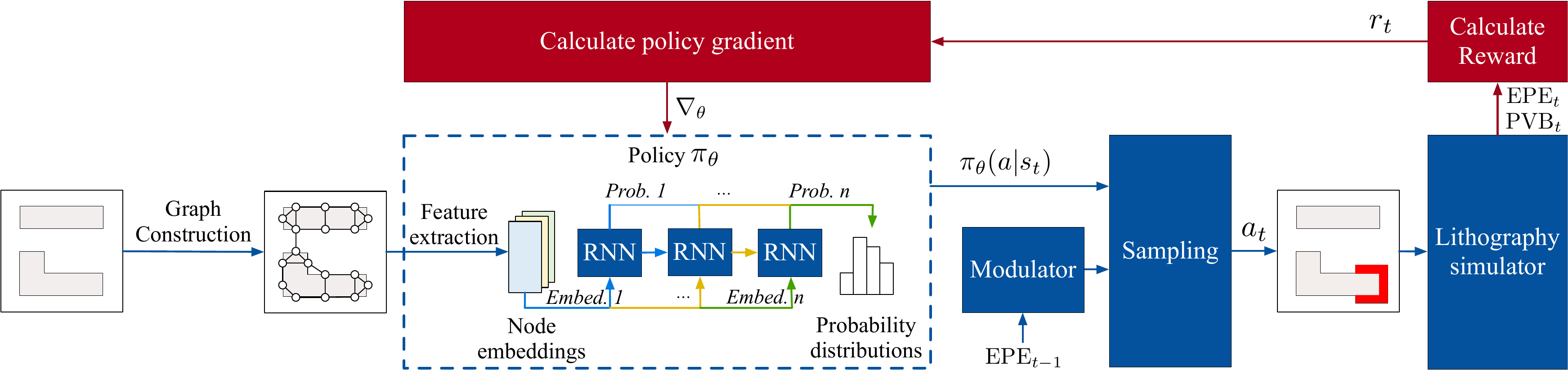}
 \caption{Overall framework of CAMO.}
 \label{fig:rl-flow}
 \end{figure*}
To build the RL framework, the key elements for RL are first illustrated, including state, action, and reward formulation.
\begin{itemize}
    \item State ($s$): The state refers to the layout geometrical information, including the up-to-date mask and the target patterns.
    \item Action ($a$): Since CAMO updates the mask by batches of segments, an action $a$ indicates the directions and strides of the movements for each segment. 
    The action space contains valid movements $\{m_1, m_2, m_3, m_4, m_5\}=\{-2nm, -1nm, 0, 1nm, 2nm\}$, where negative values and positive values correspond to inward movement and outward movement, respectively. 
    Hence, the dimension of the action space $A$ in each RL step becomes $5^n$, where $n$ is the number of segments. 
    \item Reward ($r$): Upon action, action $a$ is applied to update the mask, and a reward $r$ is generated by the environment which represents the improvement of mask quality and robustness. In this context, the reward is formulated by edge placement error (EPE) and process variation band (PV band) improvement. Similar to \cite{RLOPC-TCAD2023-Liang}, the reward $r_t$ is formulated as follows:
    \begin{equation}
     {r_t} = \frac{{\left| {EP{E_t}} \right| - \left| {EP{E_{t+1}}} \right|}}{{\left| {EP{E_{t}}} \right| + \varepsilon }}+\beta\frac{{PV{B_t} - PV{B_{t + 1}}}}{{PVB_t}},
\label{eqn:reward}
\end{equation}
    where $|EPE_t|$ and $PVB_t$ respectively indicate the EPE value and the PV band of the entire layout in step $t$, $\varepsilon$ is a small constant, and $\beta$ is a factor that adjusts the relative importance of EPE and PV band improvements.

    \item Policy ($\pi$): the policy $\pi$ refers to the decision model, which is in the form of a deep neural network in CAMO and is parameterized by $\theta$. 
\end{itemize}

In each optimizing step $t$, the local features of each segment $s_t$ are first encoded and fed into the policy network. 
The policy work then selects proper movements $a_t$ and commits to the mask, which is updated and evaluated by lithography simulation. 
The reward $r_t$ is then obtained by the EPE and PV band reduction upon the mask updating, and the policy network parameters $\theta$ are updated. 
With the updated mask, the next RL step is launched, and the iterative procedure forms the trajectory as depicted in \Cref{eqn:Markov}. 

Our RL framework is revealed in \Cref{fig:rl-flow}. 
In the training stage, given a set of target designs, the policy parameters are updated by performing epochs of mask correction and policy gradient computing.
The policy is then tested on a set of unseen designs.
Both stages consist of the following steps: (1) Formulate the segmented target pattern into a graph; (2) Generate node embeddings by aggregating the features of neighborhood segments in the mask; (3) Feed the node embeddings into the policy network, which selects an action $a_t$ for each corresponding segment; (4) Commit the specified movement to segments, and perform lithography simulation to get the reward $r_t$; (5) Move to the next iteration, repeat (2)-(4), and additionally update the policy parameters $\theta$ upon generating the reward during the training stage. 

\begin{figure}[!tb]
	\centering
	\includegraphics[width=.76\linewidth]{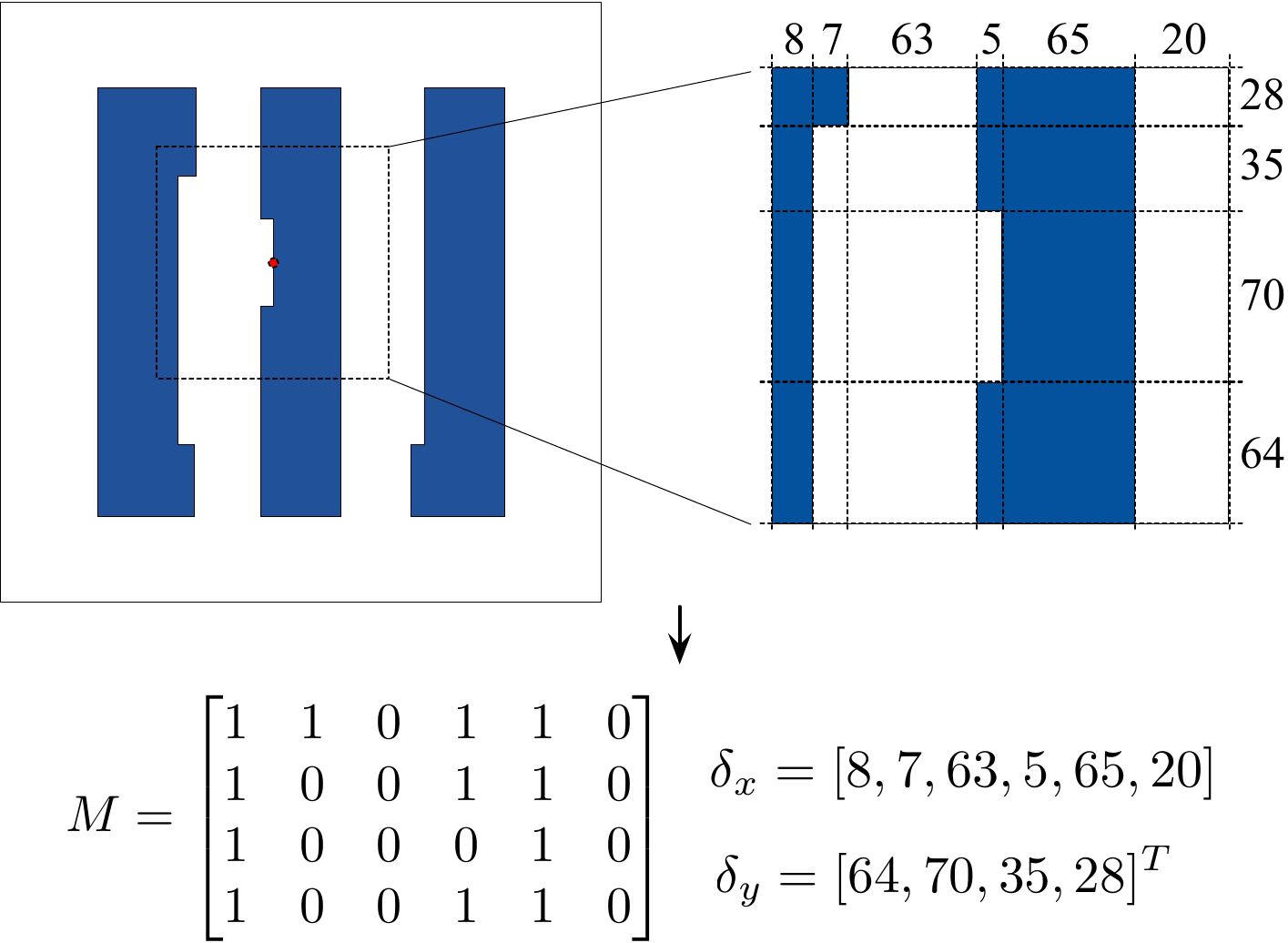}
 \caption{Illustration of squish pattern encoding. Given a control point, its neighboring geometries are encoded in $M$ and the distance information is revealed in $\delta_x$ and $\delta_x$.}
 \label{fig:SP}
 \end{figure}

\subsection{CAMO Architecture}
\label{sec:inf}
\minisection{Graph Construction.}
\label{sec:encoding}
The first step in CAMO is to construct a graph based on the given target design pattern. 
Following conventional steps in OPC, the pattern boundary fragmentation is first performed on the given pattern. 
For via patterns, the edges are regarded as segments and no fragmentation is needed. 
For metal patterns, the edges are evenly split into small segments according to the measure points such that the measure points are at the center of the segments.
The remainder is evenly absorbed by line ends. 
Taking the midpoints of each segment as the control points, the input layout is initially encoded into an undirected graph $G(\mathcal{V}, \mathcal{E})$, including node set $\mathcal{V}$ and graph edge set $\mathcal{E}$.

Each node $v\in \mathcal{V}$ corresponds to the segment, and the total number of nodes is fixed since we adopt a consistent fragmentation strategy throughout the OPC process. 
The graph edge set $E$ represents the geometry proximity among the segments, which is determined by the distance between control points. 
If the distance between two control points is closer than a threshold, an edge is added to connect the corresponding nodes. 
During the OPC process, the node features are renewed when the mask is updated, and the edge set will remain unchanged all the time.


\minisection{Correlation-aware Policy Network Structure.}
\label{sec:policy}
In CAMO, the policy structure is a neural network consisting of a GNN-based feature extractor and an RNN module, which are respectively in charge of analyzing the local features of each segment and determining movements. 

The node features are obtained by first placing a window centering at each control point and then extracting the geometry information in the window.
Specifically, since the geometries on the layout are usually sparse, directly converting the neighborhood into an image by pixels may result in redundancy in image processing.
Hence, we encode the neighborhood of each control point into squish pattern~\cite{HSD-ASPDAC2019-Yang}, as depicted in \Cref{fig:SP}. 
The window is first converted into a grid by placing scanlines at the edges of the sampled geometries, as well as two vectors $\vec{\delta_x}$ and $\vec{\delta_y}$ indicating the horizontal and vertical grid spacings by nanometers respectively. 
Subsequently, this grid is converted into a matrix $\vec{M}$, where entries corresponding to grid spaces containing geometry are marked as 1, while all others are set to 0. 
However, the policy in the form of a neural network necessitates a consistent input dimension, while the size of $\vec{M}$ varies depending on the layout geometries within the window. 
To address this inconsistency, the resulting $(\vec{M}, \vec{\delta_x}, \vec{\delta_y})$ is further converted into the adaptive squish pattern~\cite{HSD-ASPDAC2019-Yang} to formulate a tensor $\tensor{T} \in \mathbb{R}^{d_x\times d_y\times 3}$, where $d_x$ and $d_y$ denote the desired input size. 
This adaptive squish pattern is applied in previous RL-OPC~\cite{RLOPC-TCAD2023-Liang}. 
Different from \cite{RLOPC-TCAD2023-Liang}, we additionally formulate another $d_x\times d_y\times 3$ tensor in CAMO. 
The identical squish pattern encoding strategy is applied, and we place additional scanlines in both directions at the edge of the target patterns in encoding $\vec{M}$ to highlight the edge movements. 
By concatenating the two tensors, the dimension of our input node features becomes $\tensor{T}_v \in \mathbb{R}^{d_x\times d_y\times 6}$.

In the feature extractor, the node features are fused along the graph edges to capture the geometry information and produce the node embeddings. 
One example of the features fusing is GraphSAGE~\cite{GraphSage}, which involves multi-level sampling among nodes and aggregating among sampled node features. The feature fusing on node $v$ is formulated as follows:
\begin{equation}
\begin{split}
    \tensor{G}_v^{(k)}&=\operatorname{AGGREGATE}^{(k)}\left(\left\{\tensor{T}_v^{(k-1)}: u \in \mathcal{N}(v)\right\}\right),\\
    \tensor{T}_v^{(k)}&=\operatorname{COMBINE}^{(k)}\left(\tensor{T}_v^{(k-1)}, \tensor{G}_v^{(k)}\right),
\end{split}
\end{equation}
where $k$ refers to the level of sampling, $\tensor{H}_v$ refers to the node feature of node $v$, $\mathcal{N}(v)$ indicates the neighbors set of node $v$. ${\rm AGGREGATE}(\cdot)$ is the aggregation function which is usually averaging, and ${\rm COMBINE}(\cdot)$ is usually concatenating or summing.

The RNN module then sequentially processes the embeddings and recurrently records the historical contexts for future reference by maintaining a \textit{hidden state} $\vec{h}$. The forward process of a standard recurrent unit at step $t$ is formulated as follows:
\begin{equation}
\begin{split}
    {\vec{h}}^{(t)}&=\phi(\vec{U}\tensor{T}^{(t)}+\vec{W}\vec{h^{(t-1)}}+\vec{b}),\\
    \vec{o}^{(t)}&=\vec{V} \vec{h}^{(t)}+\vec{c},
\end{split}
\end{equation}
where $\tensor{T}^{(t)}$ is the embedding of the $t$-th node from the GNN, $\phi(\cdot)$ is the activation function; $\vec{U}$, $\vec{W}$ and $\vec{V}$ are the weight parameters, $\vec{b}$ and $\vec{c}$ are bias, and $\vec{o}^{(t)}$ is the output for $\tensor{T}^{(t)}$.
Through the recurrent unit, the information and movements of previous segments are considered in processing the upcoming ones. 
For each embedding, the policy outputs the estimated probability distribution of selecting the five possible movements for each segment. 


 \begin{figure}[!tb]
	\centering
	\includegraphics[width=.95\linewidth]{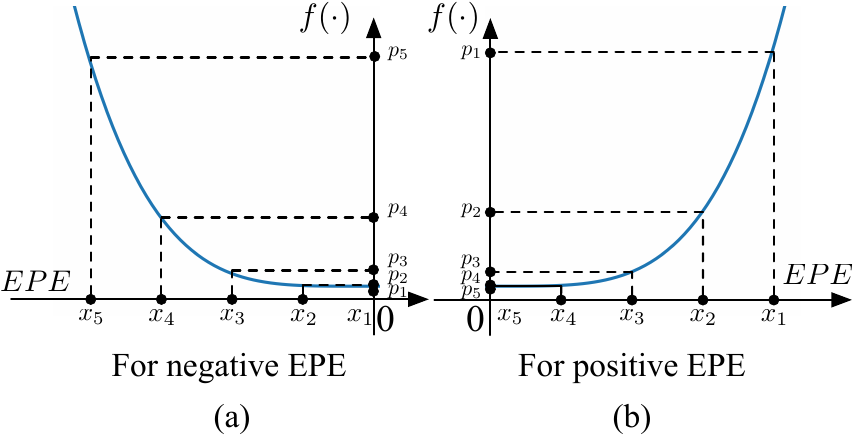}
    \caption{Illustration on the projection function. Given the EPE value of a segment, five points are evenly sampled from region $[0, EPE]$, and are then projected through $f(\cdot)$.}
 \label{fig:modu}
 \end{figure}

\minisection{OPC-Inspired Modulator.}
\label{sec:modulator}
Despite the action space is finite, it grows exponentially with the number of segments.
Especially when handling the metal layer patterns, a great diversity of patterns leads to an extremely large search space, making the policy hard to converge.
Hence, a purely data-driven approach may take tremendous time and a huge amount of data to train an agent.

In CAMO, we propose to combine the domain knowledge from the lithography principle and OPC problem to improve effectiveness and efficiency by designing a \textit{modulator}.
The modulator adjusts the likelihood of each movement being selected according to the EPE value.
It is represented by a vector $\vec{p}=[p_1, p_2, p_3, p_4, p_5]$, indicating the \textit{preference} of the five possible movements. 
Recall that EPE refers to the displacement between the printed contour and the target pattern and indicates the extent of light intensity overflowing or lacking at the target edges. 
Therefore, a modulator should satisfy the following properties: 
\begin{itemize}
\item A large inner EPE may prefer an outward movement and discourage an inner movement of the segment, and vice versa. Thus the preferences should be more distinct.
\item When EPE is small, the preferences should not be significantly biased. 
\end{itemize}
From a function's perspective, it should be flat when EPE is small and becomes sharp as EPE increases. 

Motivated by the above insights, we utilize a polynomial function $f(\cdot)=kx^n+b$, where $k$, $n$, and $b$ are positive hyper-parameters, and $n$ should be even.
Then we obtain a preference vector based on this polynomial function. 
As shown in \Cref{fig:modu}, given a signed EPE value $EPE$, we first evenly sample five points $\vec{x}=[x_1, x_2, x_3, x_4, x_5]$ from the interval $[0, EPE]$, and guarantee that $x_1 > x_2 > x_3 > x_4 > x_5$.

The sampled points serve as inputs for the projection function $f(\cdot)$ and formulate a vector $\vec{p}=f(\vec{x})$. 
Then the vector is normalized by the $softmax$ function as $\widehat{\vec{p}}=softmax(\vec{p})=[\widehat{p_1},\hdots, \widehat{p_5}]$.
After that, the vector $[\widehat{p_1},\hdots, \widehat{p_5}]$ is the modulated preference of the corresponding movements $[m_1, \hdots, m_5]$.
It can be verified that the proposed modulators satisfy the aforementioned properties.

\minisection{Decision.}
\label{sec:decision}
During the inference stage, the policy selects the movement according to the modulated probability, which can be formulated by:
\begin{equation}
    {a_t} = \mathop {\arg \max }\limits_a \widehat{\vec{p}}\odot\pi_\theta( {a|s_t}),
\label{fwd}
\end{equation}
where the modulated probability is obtained by element-wise multiplication between the output vector $\widehat{\vec{p}}$ and the probability distribution generated by the policy network.
The concatenated decision $a_t$ of $n$ segments is then committed for mask updating and evaluation.

\begin{algorithm}[!t]
    \caption{CAMO Training} 
    \label{alg1}
      \begin{algorithmic}[1]
    \Require
    $\theta_0$: initial policy parameters;
    $T$: number of training iterations;
    $M_0$: initial mask;
    \Ensure
    $\theta$: trained policy parameters
    \State Construct graph $G(\mathcal{V}, \mathcal{E})$ based on $M_0$
    \State $(EPE_0,PVB_0)\gets$ Launching lithography simulation on $M_0$
    \For{$t\gets 1 $ to $T$}   
    \State $s_t\gets$ Node feature encoding based on $M_{t-1}$ and $G(\mathcal{V}, \mathcal{E})$
    \State $\widehat{\vec{p}} \gets$ Get the modulation vector based on $EPE_{t-1}$ \Comment{Not needed in Phase 1} \label{alg:CalMod}
    \State $a_t, \pi_\theta(a_t|s_t) \gets$ Sample from $\widehat{\vec{p}} \odot \pi_\theta(a|s_t)$ \Comment{Use labeled data in Phase 1 instead of sampling} \label{alg:sample}
    \State $M_t\gets$ Take actions $a_t$ on $M_{t-1}$
    \State $EPE_t,PVB_t\gets$ Lithography simulation on $M_t$
    \State $r_t\gets$ Calculate reward by \Cref{eqn:reward}
    \State Update $\theta$ by \Cref{eqn:pgupdate} \label{alg:update}
    \EndFor
    \end{algorithmic}
\end{algorithm}

\subsection{Agent Training}
\label{sec:training}
The early stage of policy training is often time-consuming due to the RL exploration in the huge search space. 
For boosting, we adopt a two-phase training paradigm.
The training procedure is revealed in~\Cref{alg1}.

In the first phase, the policy mimics the behaviors of other OPC engines. 
For each training case, we first collect the moving trajectories of each segment within a limited number of steps by another OPC engine, e.g., Calibre. 
In step $t$, after receiving the input graph with $n$ nodes, the policy outputs the probabilities with the dimension of $n\times5$, indicating the five possible movements to be selected for $n$ segments. 
Following \cite{REINFORCE}, the update rule is formulated as follows:
\begin{equation}
\begin{split}
    \theta_{t+1}  &\leftarrow \theta_t  + \alpha {\nabla _\theta }J(\theta_t ),\\
    {\nabla _\theta }J(\theta_t ) &= {{\nabla _\theta }r(s_t, a_t)\log {\pi _\theta }\left( {a_t|s_t} \right)},
\end{split} 
\label{eqn:pgupdate}
\end{equation}
where $\alpha$ indicates the learning rate, and $\pi _\theta \left( {a|s_t} \right)$ denotes the output probability distribution when the policy $\pi_\theta$ receives $s_t$ as input. 
In this phase, the policy specifies the action $a_t$ according to the collected trajectories and updates its parameters by the output $\pi _\theta \left( {a_t|s_t} \right)$.

The next phase adopts the same policy-updating rules except for the involvement of the modulator and the selection strategy of action $a_t$. 
The training procedure is revealed in~\Cref{alg1}.
As depicted in~\Cref{alg:CalMod}, the modulation vector is calculated. 
Besides, instead of specifying $a_t$, the policy individually samples the output probability distribution with five in length for each segment as revealed in \Cref{alg:sample}. 
The $n$ decisions are then concatenated. 
After mask updating, lithography simulation, and reward calculation, the policy parameters are updated as described in~\Cref{alg:update}. 
Note that $\pi _\theta\left( {a_t|s_t} \right)$ utilized in~\Cref{eqn:pgupdate} is the original output of the policy that is independent of the modulator.




\section{Experiments}
\begin{table*}[!t]
    \small
    \centering
    \caption{OPC results comparison on via layer patterns in terms of EPE ($nm$), PV band ($nm^2$) and runtime ($s$).}
    \label{tab:viacomp}
    \resizebox{0.68\linewidth}{!}
    {
    \begin{tabular}{|cc|ccc|ccc|ccc|ccc|}
        \hline
        \multirow{2}{*}{Design} & \multirow{2}{*}{Via \#} & \multicolumn{3}{c|}{DAMO~\cite{OPC-ICCAD2020-Chen}} & \multicolumn{3}{c|}{Calibre} & \multicolumn{3}{c|}{RL-OPC~\cite{RLOPC-TCAD2023-Liang}} & \multicolumn{3}{c|}{Ours}  \\
        &                      & EPE     & PVB    & RT    & EPE      & PVB      & RT        & EPE     & PVB     & RT     & EPE  & PVB           & RT \\\hline\hline
        V1                  & 2                    &  7      & 5822   & 0.57  & 8        & 5837     & 8.11      & 6        & \textbf{5730}        & 11.12       & \textbf{1}    & 5797          & 3.18   \\
        V2                  & 2                    &  8      & 5836   & 0.56  & 8        & 5834     & 7.79      & 8        & 5813        & 8.66       & \textbf{5}    & \textbf{5734}          & 3.23   \\
        V3                  & 3                    &  14     & 8565   & 0.55  & 11       & 8587     & 8.01      & 13        & 8594        & 11.50       & \textbf{10}   & \textbf{8470}          & 4.85   \\
        V4                  & 3                    &  14     & 8621   & 0.59  & 12       & 8771     & 8.12      & 14        & 8679        & 11.79       & \textbf{10}   & \textbf{8576}          & 4.66   \\
        V5                  & 4                    &  18     & 10615  & 0.58  & 15       & 10775    & 8.18      & 16        & 10772        & 7.23       & \textbf{10}    & \textbf{10503}         & 3.28   \\
        V6                  & 4                    &  20     & 10739  & 0.58  & 15       & 10763    & 8.37      & 19        & 10659        & 11.85       & \textbf{15}   & \textbf{10507}         & 6.55   \\
        V7                  & 5                    &  28     & 12993  & 0.56  & 23       & 12615    & 8.64      & 23        & 12485        & 12.77       & \textbf{23}   & \textbf{12097}         & 11.08   \\
        V8                  & 5                    &  26     & 13047  & 0.57  & 19       & 12784    & 8.25      & 24        & 12547        & 12.42       & \textbf{19}   & \textbf{12437}         & 6.49   \\
        V9                  & 6                    &  30     & 15497  & 0.56  & 24       & 15454    & 8.70      & 26        & 15414        & 12.30       & \textbf{19}   & \textbf{15186}         & 3.45   \\
        V10                 & 6                    &  35     & 15088  & 0.58  & 27       & 15064    & 8.57      & 33        & 14588        & 12.17       & \textbf{26}   & \textbf{14556}         & 11.31   \\
        V11                 & 6                    &  39     & 15516  & 0.59  & 27       & 15782    & 8.46      & 31        & 15538        & 12.42       & \textbf{21}   & \textbf{15333}         & 3.32   \\
        V12                 & 6                    &  36     & 15424  & 0.57  & 23       & 15686    & 8.74      & 24        & 15464        & 12.59       & \textbf{23}   & \textbf{15204}         & 11.28   \\
        V13                 & 6                    &  32     & 16970  & 0.57  & 23       & 17035    & 8.43      & 39        & 17440        & 12.78       & \textbf{14}   & \textbf{16712}         & 9.70   \\\hline\hline
        Sum         & 58                               & 307     & 154733 & 7.43  & 235      & 154987   & 108.36    & 276       & 153723        & 149.6       & \textbf{196} & \textbf{151112} & 82.38   \\
        Ratio       &               & 1.57        & 1.02       & 0.10      & 1.20     & 1.03     & 1.32      & 1.41        & 1.02        & 1.96       & \textbf{1.00} & \textbf{1.00} & \textbf{1.00}  \\\hline
    \end{tabular}
    }
\end{table*}

\begin{table}[!t]
    \small
    \caption{Comparison with other OPC engines on the metal layer, in terms of EPE ($nm$), PV band ($nm^2$), and runtime ($s$).}
    \label{tab:metalcomp}
    \resizebox{1.02\linewidth}{!}
    {
    \begin{tabular}{|cc|ccc|ccc|ccc|}
        \hline
        \multirow{2}{*}{} & \multirow{2}{*}{\begin{tabular}[c]{@{}c@{}}Point\\ \#\end{tabular}} & \multicolumn{3}{c|}{Calibre} & \multicolumn{3}{c|}{RL-OPC~\cite{RLOPC-TCAD2023-Liang}} & \multicolumn{3}{c|}{Ours} \\
            &       & EPE      & PVB     & RT     & EPE     & PVB     & RT     & EPE     & PVB    & RT    \\\hline\hline
            M1      & 64       & 49      & 28728     & \textbf{8.65}      & 104        & 29390        & 16.61       & \textbf{44}     & \textbf{27795}    & 8.73     \\
            M2      & 84       & \textbf{61}      & 37386     & 8.72      & 117        & 39139        & 15.79       & 67     & \textbf{36467}    & \textbf{7.78}     \\
            M3      & 88       & 81      & \textbf{39430}     & 8.46      & 137        & 41623        & 16.88       & \textbf{59}     & 39451    & \textbf{7.95}     \\
            M4      & 100      & 89      & 45741     & \textbf{8.50}      & 252        & 46892        & 17.13       & \textbf{60}     & \textbf{44961}    & 9.41     \\
            M5      & 106      & \textbf{66}      & 47220     & \textbf{8.84}      & 336        & 47041        & 16.62       & 69     & \textbf{46582}    & 11.05     \\
            M6      & 112      & 102     & 49887     & 8.78      & 355        & 51433        & 17.57       & \textbf{78}     & \textbf{49438}    & \textbf{7.97}     \\
            M7      & 116      & 89      & 52584     & \textbf{8.92}      & 325        & 50770        & 17.82       & \textbf{83}     & \textbf{49961}    & 14.26     \\
            M8      & 24       & \textbf{20}      & 11014     & 8.51      & 32        & 10770        & 16.13       & 23     & \textbf{10928}    & \textbf{1.11}     \\
            M9      & 72       & 50      & 22531     & 8.72      & 197        & 22360        & 16.44       & \textbf{42}     & \textbf{22032}    & \textbf{8.32}     \\
            M10     & 120      & \textbf{91}      & 37546     & \textbf{8.95}      & 263        & 36368        & 16.79       & 95     & \textbf{36849}    & 11.79     \\\hline\hline
            Sum     & 886         & 698         & 372067        & \textbf{87.05}       & 2118        & 375786        & 167.78       & \textbf{620}        & \textbf{364464}       & 88.37      \\
            Ratio   &          & 1.13         & 1.02        & 0.99       & 3.42        & 1.03        & 1.90       & \textbf{1.00}        & \textbf{1.00}       & 1.00      \\\hline
    \end{tabular}
    }
\end{table}
In this section, we present experimental results on various patterns from the via layer and metal layer, and compare them with other OPC engines from academia and an industry commercial tool. 
CAMO is implemented with Python using PyTorch framework and runs on a CentOS-7 machine with an Intel i7-5930K 3.50GHz CPU and Nvidia GeForce RTX 3090 GPU. 
The Calibre-compatible lithography simulator and the corresponding Calibre OPC scripts are from an industry partner.

\subsection{Experimental Setup}
\minisection{Dataset.}
In experiments on via patterns, the layouts are adopted from~\cite{liu2020adversarial} which are $2{\mu}m \times 2{\mu}m$ clips, which contain a different number of $70nm\times70nm$ via patterns. 
The training set contains 11 clips with the number of via patterns varying from 2 to 5, and the test set contains 13 clips with 2 to 6 via patterns.
SRAFs are inserted by Calibre before CAMO launches and are included in squish pattern encoding as \cite{RLOPC-TCAD2023-Liang} did.

In experiments on metal patterns, the dataset comprises $1500nm \times 1500nm$ clips from two categories: clips from a generated GDSII layout, and clips with regular metal patterns. 
The layout is generated by OpenROAD~\cite{OPENROAD} using standard cells from NanGate 45$nm$ PDK, and the clips are randomly sampled from the metal layer.

\minisection{EPE Measure Points}
To achieve convincing comparisons, the selection of EPE measure points follows conventional strategies as in~\cite{OPC-ICCAD2013-Banerjee}.
More specifically, for a pattern in the via layer, the center of each edge corresponds to a measure point.
For a pattern in the metal layer, the EPE measure points are evenly placed on the edges along the primary direction with 60$nm$ spacing, which is consistent with the evaluation of the commercial tool we use. 
The total number of measure points is recorded in the corresponding tables. 


\minisection{CAMO Setup.}
The node features in $G(\mathcal{V}, \mathcal{E})$ are sampled with the $500nm\times500nm$ neighborhood centered at each control point, and are encoded into tensors sized $128\times128\times6$ for via layer patterns and $64\times64\times6$ for metal layer patterns. 
The threshold for determination $E$ is 250$nm$. 
Regarding the policy network structure, we adopt GraphSAGE~\cite{GraphSage} as our feature extractor. It is followed by an RNN with an input size of 256, 3 recurrent layers, a hidden state size of 64, and a $64\times5$ fully connected layer. 
The stochastic gradient descent optimizer is employed with the learning rate $\alpha$ set to $3\times 10^{-4}$. 
In calculating the modulator in \Cref{sec:modulator}, the projection function is $f(x)=0.02x^4+1$. 
In~\Cref{eqn:reward}, $\epsilon$ and $\beta$ are set to 0.1 and 1 respectively.
In the following sections, the policy $\pi$ of CAMO mimics the five-step trajectories collected from Calibre on the training set for 500 epochs.

\subsection{Experiments on Via Layer Patterns}
In experiments on via layer, we compare with ML-OPC techniques and a commercial tool Calibre~\cite{Calibre}.
We choose a state-of-the-art generative OPC model DAMO~\cite{OPC-ICCAD2020-Chen} and an RL-based approach RL-OPC~\cite{RLOPC-TCAD2023-Liang} as the baseline ML-OPC techniques.
The performance of baselines is provided by the authors of \cite{OPC-ICCAD2020-Chen,RLOPC-TCAD2023-Liang}.
Similar to RL-OPC, we also adopt the early-exit mechanism, i.e., the optimization stops when the EPE per via is less than $4nm$.
The maximum times of mask updating are set to 10, and the mask is also initialized by moving each edge outwards for $3nm$. 
DAMO is a generative model, and only a one-time inference is needed to generate a mask. Thus it is the fastest in terms of the runtime. 
However, the generative model cannot perform any exploration on the given new designs. 
Therefore the masks generated by DAMO lead to a substantially larger EPE than other methods. 
The results are revealed in \Cref{tab:viacomp}, where CAMO outperforms all three baseline methods in the EPE and PV band. 
Notably, even compared with the commercial tool, CAMO can achieve an EPE reduction by $20\%$, as well as $1.32\times$ speedup.

\subsection{Experiments on Metal Layer Patterns}
\begin{figure}[!t]
    \centering
    \hspace{-.14in}
    \subfloat[]{\includegraphics[height=1.86cm]{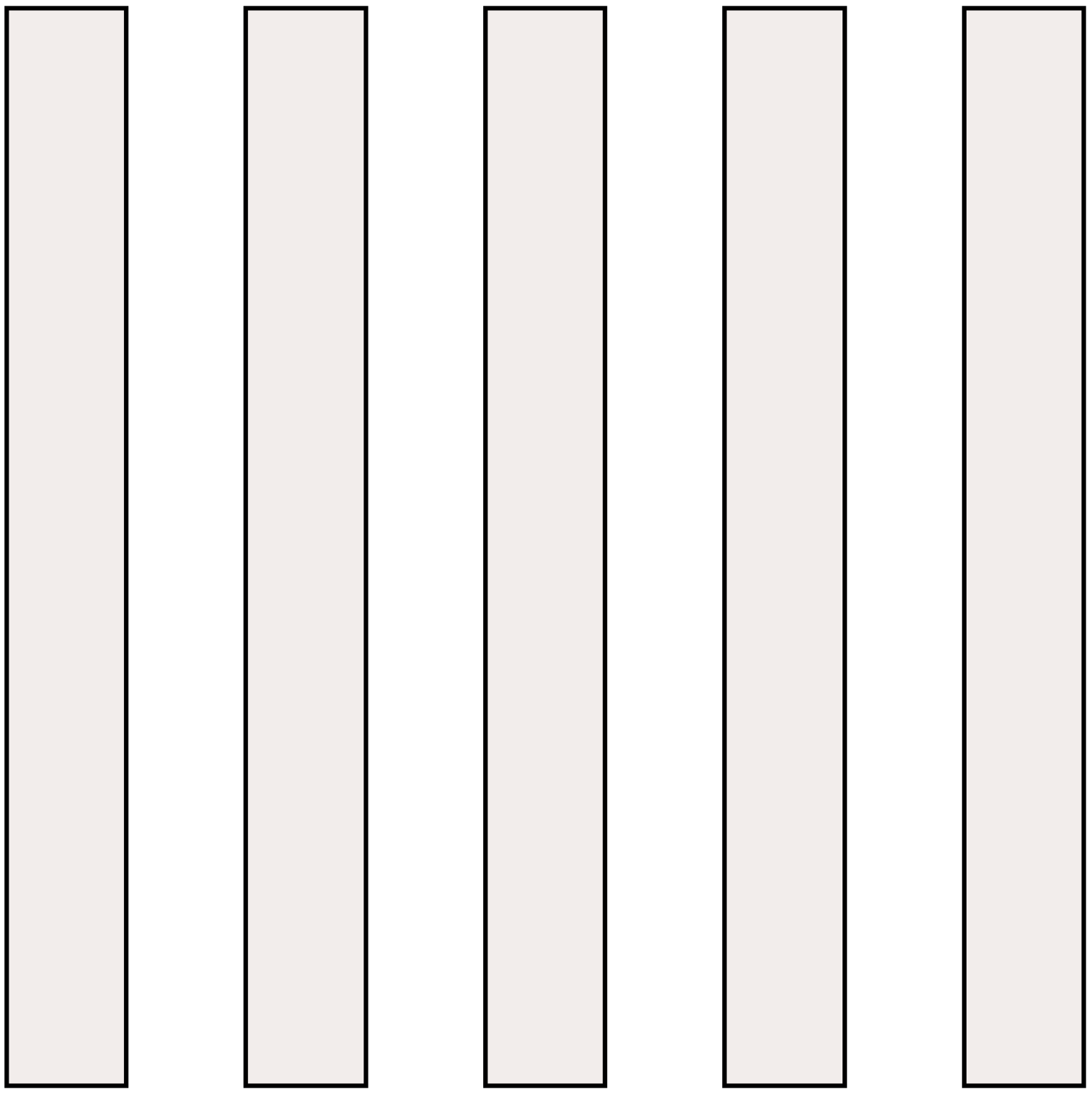}}  \hspace{.05in}
    \subfloat[]{\includegraphics[height=1.86cm]{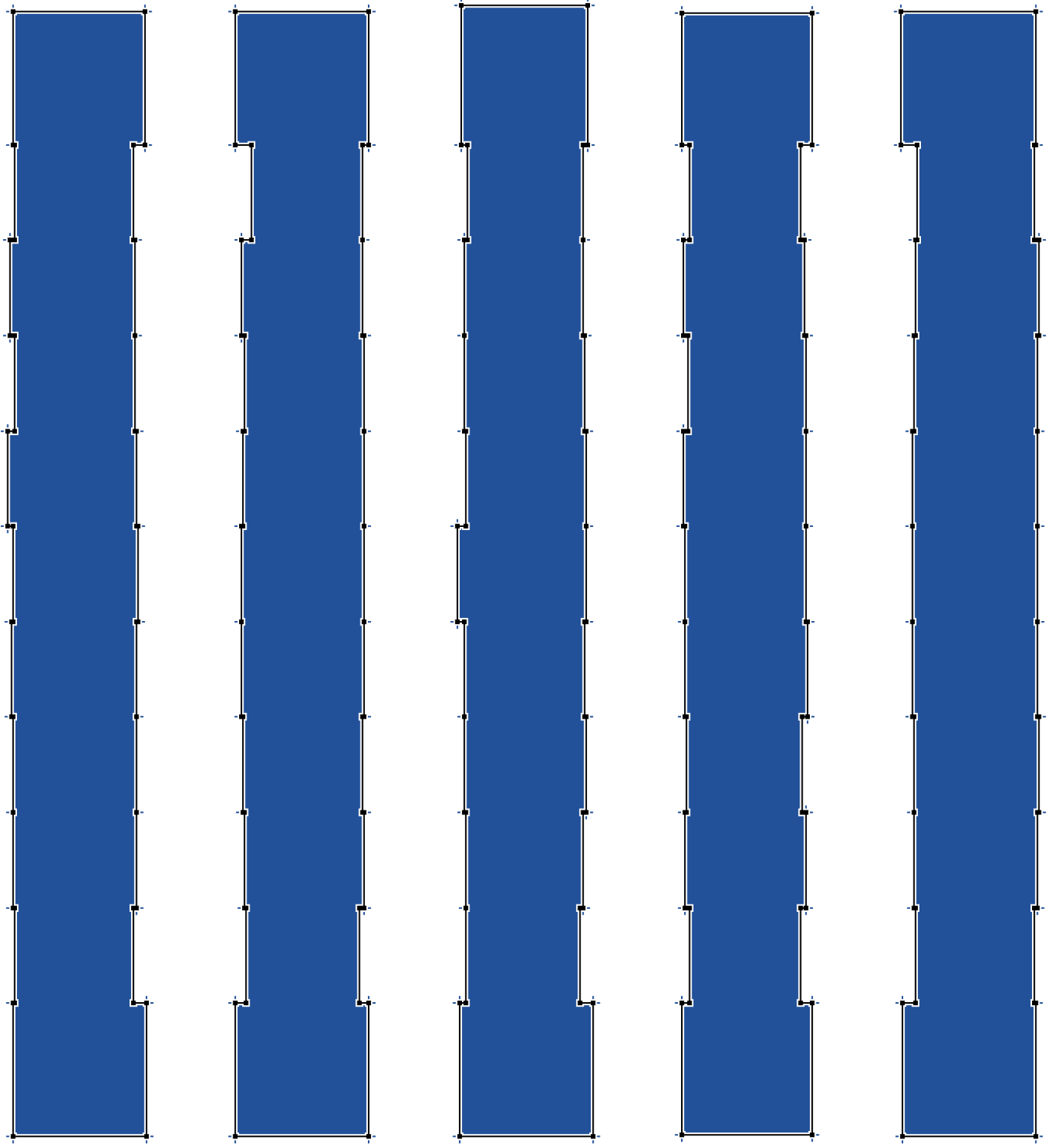}}    \hspace{.05in}
    \subfloat[]{\includegraphics[height=1.86cm]{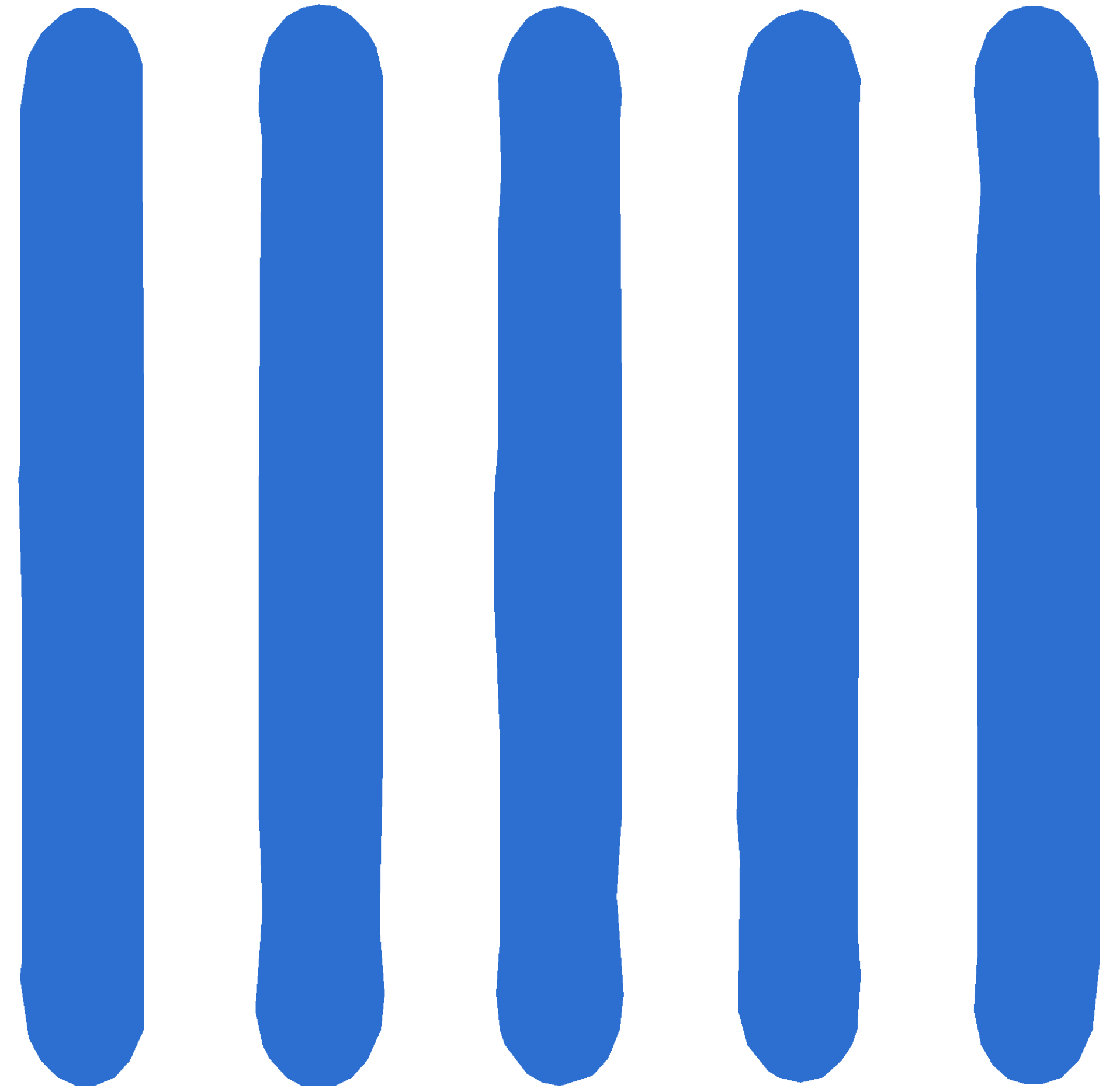}} \hspace{.05in}
    \subfloat[]{\includegraphics[height=1.86cm]{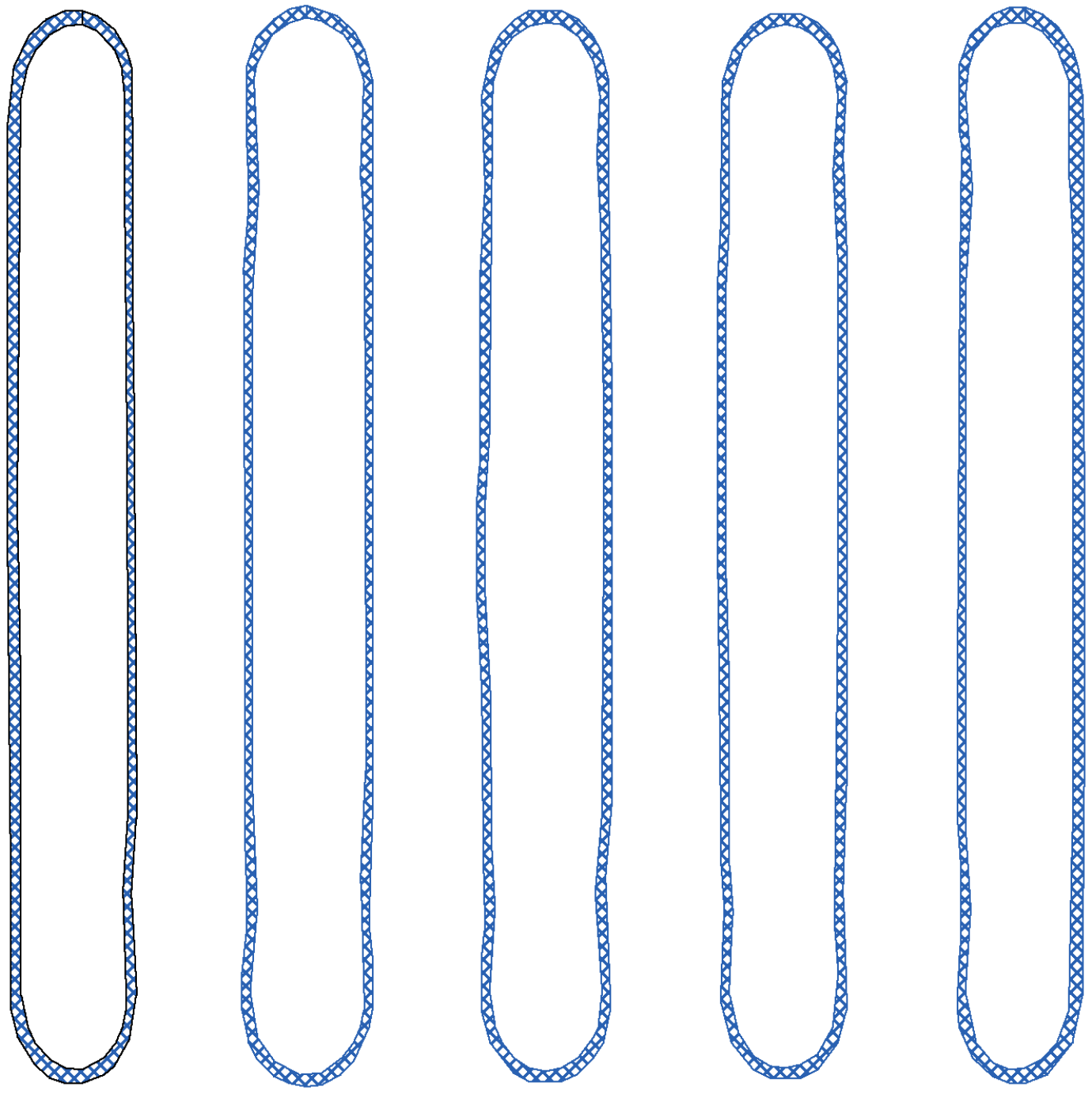}}
    \caption{An example of OPC result visualization, including (a) the target pattern, (b) the mask pattern, (c) the printed contour, and (d) the PV band.} 
    \label{visualization}

\end{figure}

For experiments on metal layer patterns, both original RL-OPC and DAMO failed to report experimental settings as well as results in their work due to the difficulty in handling more complex patterns. 
Implementing DAMO on metal layer patterns requires significantly more dataset to train and it is still hard to converge, thus DAMO cannot be used for comparison on this experiment. 
Therefore, we implement RL-OPC~\cite{RLOPC-TCAD2023-Liang} on the metal layer as the baseline. 
In our implementation, the primary settings follow the description of \cite{RLOPC-TCAD2023-Liang}.
Besides, RL-OPC and CAMO adopt identical initial masks and the early exit mechanism, where the optimization stops when the average EPE per measure point is less than $1nm$. 
The maximum number of mask updating times is 15.
In addition, we also adopt Calibre as the baseline. 
As revealed in \Cref{tab:metalcomp}, CAMO achieves a 10\% reduction in EPE and 2\% improvement in PV band compared to Calibre, as well as competitive runtime.
Our OPC result of case M10 is visualized in~\Cref{visualization} as an example.
The results of RL-OPC suggest that it is difficult to converge in this experiment, which may be due to the huge solution space upon transferring to the metal layer.

\subsection{Effectiveness of the Modulator}
\begin{figure}[!tb]
    \centering
    \definecolor{myblue}{RGB}{29,114,221}    
\definecolor{myyellow}{RGB}{255,255,191} 
\definecolor{myorange}{RGB}{244,106,18}  
\definecolor{mygray}{RGB}{102,102,102}   
\definecolor{mypink}{RGB}{252,228,215}   

\definecolor{self-purple}{RGB}{160, 135, 179}
\definecolor{self-ribbon-shallow}{RGB}{241, 223, 182}
\definecolor{self-ribbon-deep}{RGB}{221, 208, 173}
\definecolor{self-gray-shallow}{RGB}{237, 237, 237}
\definecolor{self-gray-deep}{RGB}{179, 179, 179}
\definecolor{self-orange}{RGB}{213, 165, 56}
\definecolor{self-green}{RGB}{122, 144, 89}
\definecolor{self-black}{RGB}{13, 13, 13}

\definecolor{C0}{RGB}{251,31,7}
\definecolor{C1}{RGB}{252,140,8}
\definecolor{C2}{RGB}{252,252,11}
\definecolor{C3}{RGB}{211,215,29}
\definecolor{C4}{RGB}{129,130,88}
\definecolor{C5}{RGB}{58,53,164}

\pgfplotsset{
    width=0.988\linewidth,
    height=0.616\linewidth
}

\begin{tikzpicture}[scale=0.8]
    \begin{axis}[
        name = handler,
        xlabel = {Step},
        ylabel = {EPE ($nm$)},
        axis lines = left,
        scale=0.8,
        xmin = 0,
        xmax = 15,
        ymin = 60,
        ymax = 650,
        xmajorgrids = true, ymajorgrids = true,
        legend style = {
            draw = none,
            at = {(1.0, 0.5)},
            anchor = west,
            legend columns = 1
        }
    ]

        \addlegendentry{M4 w.o. modulator}
        \addplot[
            color = USTred,
            mark = diamond*,
            mark size = 2pt,
        ] table[x={step}, y={epe}] {pgfplot/abla-mod-data/metal4-wo-mod.txt};
        
        \addlegendentry{M2 w.o. modulator}
        \addplot[
            color = USTdarkred,
            mark = *,
            mark size = 1pt,
        ] table[x={step}, y={epe}] {pgfplot/abla-mod-data/metal2-wo-mod.txt};
        
        \addlegendentry{M2 w. modulator}
        \addplot[
            color = CUHKblue,
            mark = square*,
            mark size = 1pt,
        ] table[x={step}, y={epe}] {pgfplot/abla-mod-data/metal2-w-mod.txt};

        \addlegendentry{M4 w. modulator}
        \addplot[
            color = USTgold,
            mark = o,
            mark size = 1.5pt,
        ] table[x={step}, y={epe}] {pgfplot/abla-mod-data/metal4-w-mod.txt};        
        
    \end{axis}
\end{tikzpicture}
    \caption{The EPE trajectories with / without modulator on M2 and M4 in \Cref{tab:metalcomp}.} 
    \label{plt:mod}
\end{figure}
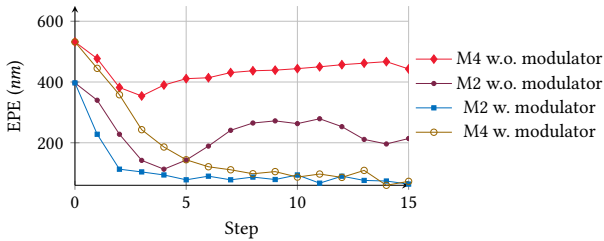

To verify the effectiveness of the proposed modulator, we compare the EPE trajectories during the optimization on two testcases with/without the modulator, as depicted in \Cref{plt:mod}. 
The fluctuating trajectories for both cases without the modulator suggest that the policy is hard to converge in the huge solution space. 
By contrast, the modulator effectively guides a more stable search in the solution space, with which the EPE curves converge and achieve at most 64$nm$ and 60$nm$ respectively for cases M2 and M4. 



\section{Conclusion}
In this paper, we propose CAMO, a spatial correlation-aware OPC system using modulated reinforcement learning. 
Different from other methods that individually decide the edge movements only based on their local features, CAMO captures the spatial correlation among neighboring segments by graph-based local feature fusing and RNN-based sequential decision. 
Besides, we design a modulator inspired by the intuition of OPC, which improves the effectiveness and efficiency of CAMO.
Finally, CAMO outperforms state-of-the-art OPC engines from academic approaches and a commercial toolkit.
\section*{Acknowledgments}
This work is supported in part by the National Natural Science Foundation of China (No. 62204066, No. 62202190), Guangzhou Municipal Science and Technology Project (Municipal Key Laboratory Construction Project, Grant No.2023A03J0013), The Research Grants Council of Hong Kong SAR (Project No.~CUHK14208021), and Hubei National Science Foundation (No. 2023AFB237).

{
\bibliographystyle{IEEEtran}
\bibliography{ref/Top-sim,ref/Others,ref/DFM,ref/HSD}
}
\end{document}